\documentclass[a4paper,UKenglish,cleveref, autoref, thm-restate]{lipics-v2021}
\nolinenumbers

\usepackage[utf8]{inputenc} % allow utf-8 input
\usepackage[T1]{fontenc}    % use 8-bit T1 fonts
\usepackage{hyperref}       % hyperlinks
\usepackage{url}  % simple URL typesetting
\usepackage{booktabs}       % professional-quality tables
\usepackage{amsfonts}       % blackboard math symbols
\usepackage{nicefrac}       % compact symbols for 1/2, etc.
\usepackage{microtype}      % microtypography
\usepackage{lipsum}		% Can be removed after putting your text content
\usepackage{graphicx}
\usepackage{doi}
\usepackage{algpseudocode}
\usepackage[linesnumbered,ruled,vlined]{algorithm2e}

\usepackage{tabularx, multirow, graphicx}
\usepackage{multirow}
\usepackage[table,xcdraw]{xcolor}
\newcolumntype{C}{>{\centering\arraybackslash}X}
\newcolumntype{P}[1]{>{\centering\arraybackslash}p{#1}}

\usepackage{siunitx}
\usepackage{float}
\usepackage{amsmath,amssymb}

\usepackage{bm}
\usepackage{caption}
\usepackage{subcaption}
\usepackage{xcolor}
\usepackage{array}
\usepackage{mathtools}
\usepackage{tabstackengine}
\usepackage{transparent}
\usepackage{cancel}
\usepackage{soul}
\usepackage[renew-dots,renew-matrix]{nicematrix}

\newenvironment{smallarray}[1]
 {\null\,\vcenter\bgroup\scriptsize
  \arraycolsep=.13885em
  \hbox\bgroup$\array{@{}#1@{}}}
 {\endarray$\egroup\egroup\,\null}

\makeatletter
\newcommand{\colormathbox}[3][\mathord]{%
  #1{%
    \setlength{\fboxsep}{1pt}%
    \mathpalette\color@mathbox{{#2}{#3}}%
  }%
}
\newcommand{\color@mathbox}[2]{%
  \color@@mathbox#1#2%
}
\newcommand{\color@@mathbox}[3]{%
  \colorbox{#2}{$#1\m@th#3$}%
}
\makeatother

\title{Fast Matrix Multiplication Without Tears: \\ A Constraint Programming Approach}

\author{Arnaud Deza\footnote{\label{note1}These authors contributed equally.}}{Department of Mechanical and Industrial Engineering, University of Toronto, Ontario, Canada}{arnaud.deza@mail.utoronto.ca}{}{}

\author{Chang Liu\protect\footnotemark[1]}{Department of Mechanical and Industrial Engineering, University of Toronto, Ontario, Canada}{changy.liu@mail.utoronto.ca}{}{}%{https://orcid.org/0000-0003-2808-6056}

\author{Pashootan Vaezipoor}{Department of Computer Science, University of Toronto, Ontario, Canada}{pashootan@cs.toronto.edu}{}{}%{https://orcid.org/0000-0003-2808-6056}

\author{Elias B. Khalil}{Department of Mechanical and Industrial Engineering, University of Toronto, Ontario, Canada}{khalil@mie.utoronto.ca}{}{}%{https://orcid.org/0000-0003-2808-6056}

\authorrunning{A. Deza, C. Liu, E.B. Khalil, P. Vaezipoor} 

\Copyright{Arnaud Deza, Chang Liu, Elias B. Khalil, Pashootan Vaezipoor} %TODO mandatory, please use full first names. LIPIcs license is "CC-BY";  http://creativecommons.org/licenses/by/3.0/


\supplement{}
\supplementdetails[linktext={https://github.com/khalil-research/Matrix-\\Mult-CP}, subcategory={Source Code}, cite={}]{Software}{https://github.com/khalil-research/Matrix-Mult-CP}

\EventEditors{Roland H. C. Yap}
\EventNoEds{1}
\EventLongTitle{29th International Conference on Principles and Practice of Constraint Programming (CP 2023)}
\EventShortTitle{CP 2023}
\EventAcronym{CP}
\EventYear{2023}
\EventDate{August 27--31, 2023}
\EventLocation{Toronto, Canada}
\EventLogo{}
\SeriesVolume{280}
\ArticleNo{26}
\begin{document}
\maketitle

\begin{abstract}
It is known that the multiplication of an $N \times M$ matrix with an $M \times P$ matrix can be performed using fewer multiplications than what the naive $NMP$ approach suggests. The most famous instance of this is Strassen's algorithm for multiplying $2\times 2$ matrices in 7 instead of 8 multiplications. This gives rise to the constraint satisfaction problem of fast matrix multiplication, where a set of $R < NMP$ multiplication terms must be chosen and combined such that they satisfy correctness constraints on the output matrix. Despite its highly combinatorial nature, this problem has not been exhaustively examined from that perspective, as evidenced for example by the recent deep reinforcement learning approach of AlphaTensor. In this work, we propose a simple yet novel Constraint Programming approach to find algorithms for fast matrix multiplication or provide proof of infeasibility otherwise. We propose a set of symmetry-breaking constraints and valid inequalities that are particularly helpful in proving infeasibility. On the feasible side, we find that exploiting solver performance variability in conjunction with a sparsity-based problem decomposition enables finding solutions for larger (feasible) instances of fast matrix multiplication. Our experimental results using CP Optimizer demonstrate that we can find fast matrix multiplication algorithms for matrices up to $3\times 3$ with $R=23$ in a short amount of time. %Our code is available on GitHub\footnote{\url{https://github.com/khalil-research/Matrix-Mult-CP}}.%Proving infeasibility is more difficult but benefits tremendously from symmetry-breaking.
\end{abstract}

% \vspace{-0.6cm}
%%%%%%%%%%%%%%%%%%%%%%%%%%%%%%%%%%%%%%%%%%%%%%%%%%%%%%%%%
\section{Introduction} \label{sec:intro}
%%%%%%%%%%%%%%%%%%%%%%%%%%%%%%%%%%%%%%%%%%%%%%%%%%%%%%%%%
Matrix multiplication is a fundamental operation in linear algebra with applications in virtually every computational domain. As a result, extensive research has been dedicated to the development of faster matrix multiplication algorithms.

The elementary way of multiplying two $N \times N$ matrices requires $N^3$ multiplications. For example, multiplying two $2 \times 2$ matrices naively requires a total of $2^3 = 8$ multiplications. In 1969, Strassen~\cite{strassen1969gaussian} constructed an algorithm that finds the product of two $2 \times 2$ matrices in only 7 multiplications. This discovery has had significant implications as it opened up the door for potentially faster algorithms for large-scale matrix or tensor computations. Strassen's algorithm has later been proved to be both canonical~\cite{brockett1973optimal} (no smaller rank exists) and essentially unique~\cite{de1978varieties} (all other solutions of the same rank are equivalent up to symmetry). 

Currently, the best-known algorithm for multiplying $3\times 3$ matrices requires $R = 23$ multiplications, compared to the naive elementary method that requires 27 multiplications. A known theoretical lower bound of $R = 19$ exists~\cite{blaser2003complexity}, however, it remains unclear whether $19 \leq R \leq 22$ is truly attainable. This is a testament to the difficulty of the \emph{fast matrix multiplication} (FMM) problem, which has been intractable for existing methods even for tiny matrices.

In the literature, the general approach to finding FMM algorithms starts by representing matrix multiplication as a tensor operation using the multiplication tensor $T_N$ followed by finding exact or approximate low-rank decompositions that represent $T_N$. The factor matrices that are used in the low-rank decomposition encode FMM algorithms. A rank-7 decomposition (i.e., a multiplication algorithm that uses 7 multiplication operations) of a $2 \times 2$ matrix multiplication using Strassen's algorithm is shown in Figure~\ref{fig:tensor}. Existing methods for finding such factor matrices have several limitations. The most successful and common methods include local search~\cite{Smirnov2013TheBC} techniques for low-rank approximation, which cannot guarantee optimality. A more recent successful approach~\cite{fawzi2022discovering} searches for low-rank decomposition using \emph{reinforcement learning} (RL) and was successful in finding faster algorithms for $N=4$. However, this method is not exhaustive and hence cannot prove the infeasibility of a given rank.

In this work, we propose a novel approach to finding FMM algorithms by formulating the tensor decomposition problem, for the first time, as a \emph{constraint satisfaction problem} (CSP) that is solved using \emph{Constraint Programming} (CP). We believe that this is a very natural formulation of this highly combinatorial problem. CP is advantageous for FMM in that it is a flexible framework that can bring to bear a wide range of search and logical inference techniques that have been developed over the last few decades. It provides the ability to prove infeasibility when it is not possible to multiply two matrices using a given number of multiplications. 

Besides a base CP formulation for FMM, we propose a set of symmetry-breaking constraints and valid inequalities that are useful for infeasibility proofs. On the feasible side, we show that ``performance variability'' w.r.t. solver random seeds can be exploited in conjunction with a sparsity-based decomposition of FMM for faster solving. Our experimental results, while limited to matrices of size up to $3\times 3$, demonstrate the effectiveness of the aforementioned constraints and techniques. The CP approach to FMM is uniquely positioned to close open questions such as whether it is possible to multiply two $3\times 3$ matrices in 19 to 22 multiplications. While we do not yet resolve this or other open questions, our work opens up the potential for further enhancements to the CP formulation and search such as customized branching strategies and CP-based heuristics.

%%%%%%%%%%%%%%%%%%%%%%%%%%%%%%%%%%%%%%%%%%%%%%%%%%%%%%%%%
\section{Fast Matrix Multiplication: Problem Statement}
%%%%%%%%%%%%%%%%%%%%%%%%%%%%%%%%%%%%%%%%%%%%%%%%%%%%%%%%%
The multiplication of two matrices $A$ and $B$ of sizes $N \times M$ and $M \times P$, respectively, results in a product matrix $C$ of size $N \times P$. This operation can be represented by a binary third-order tensor $T_{NMP}$ ($T_{N}$ for square matrices $A$ and $B$ of size $N\times N$). An entry $T_{i,j,k}$ of this tensor is equal to 1 if and only if the $k^{\text{th}}$ entry in the output matrix $C$ uses the scalar product of the $i^{\text{th}}$ entry of $A$ and the $j^{\text{th}}$ entry of $B$. Here, $i$, $j$, and $k$ are indices of a matrix entry starting with 1 in the first row and column; and proceeding entry by entry, left to right, top to bottom. For example, for $N=M=P=2$, it must be that $T_{2,3,1}=1$ since the first entry of $C$, $c_1$, is equal to $a_1 b_1 + \mathbf{a_2 b_3}$. Similarly, $T_{1,2,1}=0$ must hold since $a_1 b_2$ is not part of $c_1$.  Figures~\ref{fig:2by2} and~\ref{fig:t2} show a complete example of the indexing and tensor representation.

The FMM problem for a given tensor $T_{NMP}$, rank $R\in\mathbb{Z}^{+}$, and field $\mathbb{F}$ (e.g., $\mathbb{F}=\{-1,0,1\}$) asks: can each entry $T_{i,j,k}$ of $T_{NMP}$ be expressed as the sum of exactly $R$ trilinear terms involving the \textit{factor matrices} $U\in\mathbb{F}^{N\cdot M\times R}$, $V\in\mathbb{F}^{M\cdot P\times R}$, and $W\in\mathbb{F}^{N\cdot P\times R}$, as follows:
% One can decompose these product tensors as a sum of rank-1 terms denoted by factor matrices  as follows: 
\begin{equation*}
    T_{i,j,k} = \sum_{r = 1}^{R} U_{i,r} \cdot V_{j,r} \cdot W_{k,r} \;\;\;\;\; \forall i \in \{1, \dots, N \cdot M\}, j \in \{1, \dots, M \cdot P\}, k \in \{1, \dots, N \cdot P\} 
\end{equation*}
Note that we use the notation $\mathbb{F}^{L\times Q}$ to refer to the set of matrices of dimension $L\times Q$ and entries in $\mathbb{F}$. The CSP is to find factor matrices with entries in $\mathbb{F}$ that produce the tensor $T_{NMP}$ for a given rank $R$. 

This decomposition is also referred to as the polyadic decomposition and its associated rank is the minimal $R$ needed. The rank can be interpreted as the number of multiplications required to compute the product. For example, for $2\times 2$ matrices, the rank of the decomposition using Strassen's algorithm is 7. Figure~\ref{fig:tensor} walks through an example of the low-rank decomposition of a $2 \times 2$ matrix multiplication using Strassen's algorithm. The matrix multiplication of the two $2 \times 2$ matrices can be seen in Figure~\ref{fig:2by2}, its associated tensor representation $T_N$ in Figure~\ref{fig:t2}, the low-rank decomposition in Figure~\ref{fig:decomp}, and the factor matrices $U$, $V$, and $W$ in Figure~\ref{fig:factor}. 

\begin{figure}[htbp]
    % \centering
    \begin{subfigure}[b]{\textwidth}
        \begin{align*}
            \left(\begin{array}{cc}
            \cellcolor{blue!20}{c_1} & c_2 \\
            c_3 & c_4
            \end{array} \right) &=
            \left(\begin{array}{cc}
            \cellcolor{red!20}{a_1} & \cellcolor{yellow!50}{a_2} \\
            a_3 & a_4
            \end{array} \right)\cdot
            \left(\begin{array}{cc}
            \cellcolor{red!20}{b_1} & b_2 \\
            \cellcolor{yellow!50}{b_3} & b_4
            \end{array} \right)
        \end{align*}
    \vspace{-3mm}
    \caption{Multiplication of two $2 \times 2$ matrices. We highlight the term $c_1 = a_1b_1 + a_2b_3$.}
    \label{fig:2by2}
    \end{subfigure}

    \begin{subfigure}[b]{\textwidth}
    \begin{align*}
        \colorbox{blue!20}{$T_{:,:,1}$} = \begin{psmallmatrix}
          \colormathbox{red!20}{\mathbf{1}} & 0 & 0 & 0\\
          0 & 0 & \colormathbox{yellow!50}{\mathbf{1}} & 0\\
          0 & 0 & 0 & 0\\
          0 & 0 & 0 & 0
        \end{psmallmatrix} \;\;\;\; 
        T_{:,:,2} = \begin{psmallmatrix}
          0 & \mathbf{1} & 0 & 0\\
          0 & 0 & 0 & \mathbf{1}\\
          0 & 0 & 0 & 0\\
          0 & 0 & 0 & 0
        \end{psmallmatrix} \;\; \;\;
        T_{:,:,3} = \begin{psmallmatrix}
          0 & 0 & 0 & 0\\
          0 & 0 & 0 & 0\\
          \mathbf{1} & 0 & 0 & 0\\
          0 & 0 & \mathbf{1} & 0
        \end{psmallmatrix} \;\;\;\; 
         T_{:,:,4} = \begin{psmallmatrix}
          0 & 0 & 0 & 0\\
          0 & 0 & 0 & 0\\
          0 & \mathbf{1} & 0 & 0\\
          0 & 0 & 0 & \mathbf{1}\\
        \end{psmallmatrix}
    \end{align*}
    \vspace{-3mm}
    \caption{Tensor representation of the $2 \times 2$ matrix multiplication operation. $T_{:,:,1}$ represents $c_1$, the entry $T_{2,3,1}$ (in yellow) is set to $1$  because the product $a_2 b_3$ is required to compute $c_1$ (similarly for $T_{1,1,1}$ in red).}
        \label{fig:t2}
    \end{subfigure}

    \begin{subfigure}[c]{\textwidth}
        \begin{align*}
            \setlength\fboxsep{0pt}
            \colorbox{green!20}{\strut $m_1$} &= \colorbox{green!20}{$(a_1 + a_4)(b_1 + b_4)$} &  m_5 &= (a_1 + a_2)(b_4) \\
            m_2 &= (a_3 + a_4)(b_1) &  m_6 &= (a_3 - a_1)(b_1 + b_2)\\ %
            m_3 &= (a_1)(b_2 - b_4) & m_7 &= (a_2 - a_4)(b_3 + b_4)\\
            m_4 &= (a_4)(b_3 - b_1) 
        \end{align*}
        \begin{align*}
            \colorbox{blue!20}{$c_1$} &= \colorbox{green!20}{$m_1$} + m_4 - m_5 + m_7 \\
            &= \colorbox{green!20}{$(a_1 + a_4)(b_1 + b_4)$} + (a_4)(b_3 - b_1) - (a_1 + a_2)(b_4) + (a_2 - a_4)(b_3 + b_4) \\
            &= a_1 b_1 + \cancel{a_1b_4} + \cancel{a_4b_1} + \cancel{a_4b_4} + \cancel{a_4b_3} - \cancel{a_4b_1} - \cancel{a_1b_4} - \cancel{a_2b_4} + a_2b_3 + \cancel{a_2b_4} - \cancel{a_4b_3} - \cancel{a_4b_4} \\
            &=\colorbox{red!20}{$a_1b_1$} + \colorbox{yellow!50}{$a_2b_3$} \\
            c_2 &= m_3 + m_5 \\
            c_3 &= m_2 + m_4 \\
            c_4 &= m_1 - m_2 + m_3 + m_6
        \end{align*}
    \vspace{-5mm}
    \caption{A low-rank decomposition of the $2 \times 2$ matrix multiplication using Strassen's algorithm. The $m$ terms are the multiplication terms and the $c$ terms represent the entries in the product matrix. Here $c_1 = m_1 + m_4 - m_5 + m_7$ gives $c_1 = a_1b_1 + a_2b_3$ after expansion.}
    \label{fig:decomp}
    \end{subfigure}

    \newlength\mylen
    \settowidth\mylen{$m_6$}
    
    \begin{subfigure}[c]{\textwidth}
        \begin{align*}
             & \begin{array}{p{0.5mm}p{4.5mm}p{4.5mm}p{4.5mm}p{4.5mm}p{4.5mm}p{4.5mm}p{4.5mm}}
            \! & \cellcolor{green!20}{$\mathit{\!m_1}$} & $\mathit{m_2}$ & $\mathit{m_3}$ & $\mathit{m_4}$ & $\mathit{m_5}$ & $\mathit{m_6}$ & $\mathit{m_7}$
            \end{array} \\
            % U = & \left( \begin{array}{p{3mm}p{3mm}p{3mm}p{3mm}p{3mm}p{3mm}p{3mm}}
            %   \cellcolor{green!20}{1}&0&1&0&1&-1&0\\
            %   \cellcolor{green!20}{0}&0&0&0&1&0&1\\
            %   \cellcolor{green!20}{0}&1&0&0&0&1&0\\
            %   \cellcolor{green!20}{1}&1&0&1&0&0&-1
            %     \end{array} \right) 
            %     \begin{array}{c}
            % \mathit{a_1}\\
            % \mathit{a_2}\\
            % \mathit{a_3}\\
            % \mathit{a_4}
            % \end{array}
            %     \\
            % U = & \left( \begin{array}{rrrrrrr}
            U = & \left( \begin{array}{*{7}{wr{\mylen}}}
              \cellcolor{green!20}{1}&0&1&0&1&$-1$&0\\
              \cellcolor{green!20}{0}&0&0&0&1&0&1\\
              \cellcolor{green!20}{0}&1&0&0&0&1&0\\
              \cellcolor{green!20}{1}&1&0&1&0&0&$-1$
                \end{array} \right) 
                \begin{array}{c}
            \mathit{a_1}\\
            \mathit{a_2}\\
            \mathit{a_3}\\
            \mathit{a_4}
            \end{array}
                \\
            % V = &\left( \begin{array}{p{3mm}p{3mm}p{3mm}p{3mm}p{3mm}p{3mm}p{3mm}}
            %   \cellcolor{green!20}1&1&0&-1&0&1&0\\
            %   \cellcolor{green!20}0&0&1&0&0&1&0\\
            %   \cellcolor{green!20}0&0&0&1&0&0&1\\
            %   \cellcolor{green!20}1&0&-1&0&1&0&1
            %     \end{array} \right) 
            %     \begin{array}{c}
            % \mathit{b_1}\\
            % \mathit{b_2}\\
            % \mathit{b_3}\\
            % \mathit{b_4}
            % \end{array}
            %     \\
            V = & \left( \begin{array}{*{7}{wr{\mylen}}}
              \cellcolor{green!20}1&1&0&$-1$&0&1&0\\
              \cellcolor{green!20}0&0&1&0&0&1&0\\
              \cellcolor{green!20}0&0&0&1&0&0&1\\
              \cellcolor{green!20}1&0&$-1$&0&1&0&1
                \end{array} \right) 
                \begin{array}{c}
            \mathit{b_1}\\
            \mathit{b_2}\\
            \mathit{b_3}\\
            \mathit{b_4}
            \end{array}
                \\
            % W = &\left( \begin{array}{p{3mm}p{3mm}p{3mm}p{3mm}p{3mm}p{3mm}p{3mm}} 
            %   \cellcolor{blue!20}1 &\cellcolor{blue!20}0 &\cellcolor{blue!20}0 &\cellcolor{blue!20}1 &\cellcolor{blue!20}-1 &\cellcolor{blue!20}0 &\cellcolor{blue!20}1\\
            %   0&0&1&0&1&0&0\\
            %   0&1&0&1&0&0&0\\
            %   1&-1&1&0&0&1&0 
            %     \end{array} \right) 
            %     \begin{array}{c}
            % \mathit{c_1}\\
            % \mathit{c_2}\\
            % \mathit{c_3}\\
            % \mathit{c_4}
          W = & \left( \begin{array}{*{7}{wr{\mylen}}}
              \cellcolor{blue!20}1 &\cellcolor{blue!20}0 &\cellcolor{blue!20}0 &\cellcolor{blue!20}1 &\cellcolor{blue!20}$-1$ &\cellcolor{blue!20}0 &\cellcolor{blue!20}1\\
              0&0&1&0&1&0&0\\
              0&1&0&1&0&0&0\\
              1&$-1$&1&0&0&1&0 
                \end{array} \right) 
                \begin{array}{c}
            \mathit{c_1}\\
            \mathit{c_2}\\
            \mathit{c_3}\\
            \mathit{c_4}
        \end{array}
        \end{align*}
    \vspace{-3mm}
    \caption{The factor matrices $U$, $V$, and $W$ for Strassen's algorithm. The columns in $U$ and $V$ represent the coefficient of the $a$ and $b$ terms in each $m$. Each row in $W$ represents the coefficient of the $m$ terms in one $c$ term.}
    \label{fig:factor}
    \end{subfigure}
    
    \caption{A low-rank decomposition of a $2 \times 2$ matrix multiplication using Strassen's algorithm.}
    \label{fig:tensor}
\end{figure}

%%%%%%%%%%%%%%%%%%%%%%%%%%%%%%%%%%%%%%%%%%%%%%%%%%%%%%%%%
\section{Related Work}
%%%%%%%%%%%%%%%%%%%%%%%%%%%%%%%%%%%%%%%%%%%%%%%%%%%%%%%%%
Since Strassen's discovery~\cite{strassen1969gaussian}, there has been substantial research on finding faster algorithms for matrix multiplication. Mathematicians have discovered such algorithms manually over the years for a variety of matrix dimensions and ranks. In this section, however, we will focus on automated methods for discovering such algorithms and briefly discuss some of the existing methods. A recent survey on the topic can be found in~\cite{gs005}.

\subsection{Continuous Local Search Methods}
The most common approach in the literature to compute the factor matrices $U$, $V$, and $W$ is to use (heuristic, continuous) local search methods for low-rank tensor decomposition. The state-of-the-art local method~\cite{Smirnov2013TheBC} uses alternating least squares with regularization. This method has been the most successful in finding fast algorithms whilst remaining computationally tractable and has been scaled up to $N=M=P=4$, $R=49$\footnote{Note that this particular result is not very useful as an $R=49$ solution can be obtained by applying Strassen's $R=7$ algorithm for $2\times 2$ matrices on the four  $2\times 2$ blocks of the  $4\times 4$ matrices.}.  However, this approach has limitations which include getting stuck at local minima, facing ill-conditioned linear least-squares problems, and solutions being only adequate up to machine precision. Additionally, these methods are not exhaustive and hence cannot be used to provide a proof of infeasibility for a given rank $R$.

\subsection{AlphaTensor}
More recently, DeepMind released AlphaTensor~\cite{fawzi2022discovering}, a deep RL method that searches this large combinatorial space by playing a single-player game, the TensorGame, formulated as a \emph{Markov decision process} (MDP). At every step $t$ of this MDP, the state is characterized by a tensor $S_t$ which is initially set to the target multiplication tensor, i.e., $S_0 = T_N$. An action $a_t$ at iteration $t$ corresponds to the player selecting a triplet of vectors $(u^{(t)},v^{(t)},w^{(t)})$ which in turn will provide the next state $S_{t} = S_{t-1} - u^{(t)} \otimes v^{(t)} \otimes w^{(t)}$ where $\otimes$ denotes the outer tensor product. The goal of the player is to reach the zero tensor $S_t = \mathbf{0}$ in the fewest number of steps possible. This is done by providing a reward of $-1$ to the player after every non-terminal state whereas a large negative reward $-\gamma(S_{R_{\textrm{limit}}})$ is given to the player if the number of steps $R_{\textrm{limit}}$ is met, where $\gamma(S_{R_{\textrm{limit}}})$ upper bounds the rank of the tensor at iteration $R_{\textrm{limit}}$. If the agent successfully reaches the zero tensor, the sequence of actions taken constitutes a valid low-rank decomposition of $T_N$, and hence an FMM algorithm is found with the rank $R$ corresponding to the number of steps taken by the agent.

This approach is the first to directly incorporate learning into the search which resulted in the discovery of new minimal ranks for certain non-trivial cases. The largest case tackled by this method is $N=M=P=5$, $R=98$. The sole focus of this purely heuristic method is to find lower ranks than currently best-known ranks but it cannot prove the infeasibility of a given rank. Additionally, rather complex architectures and multiple training phases were required for successful learning. It is worth noting that AlphaTensor was trained for one week on 64 Tensor Processing Units (TPUs), Google's proprietary chip. The paper~\cite{fawzi2022discovering} does not provide any estimates of the amount of computation required to produce the reported results, namely how long the trained ``agent'' must be run to discover FMM algorithms. Our CP runs use much fewer resources while leveraging thread parallelism in the CP solver on readily-available CPU machines. 

\subsection{Integer Programming}
The work that is the most related to our approach tackles this problem through a \emph{mixed-integer linear program} (MILP) formulation in an unpublished technical report~\cite{sorber2017mixed}. The goal of this methodology is to linearize the trilinear products in the low-rank decomposition of $T_N$ to a MILP that aims to 1) maximize the sparsity of the integer decision variables representing factor matrices $U$, $V$, and $W$ and 2) minimize the reconstruction loss (L1 norm) from the input $T_N$ and the multiplication tensor attained by the decision variables representing factor matrices. The report~\cite{sorber2017mixed}  focuses solely on presenting the MILP formulation for square matrices but does not include any computational experiments.  However, the MILP formulations for $N\in\{2,3\}$ are benchmark problems in MIPLIB 2017~\cite{gleixner2021miplib}\footnote{See \url{https://miplib.zib.de/instance_details_fastxgemm-n3r21s3t6.html} for example.}. The linearization of the trilinear products likely leads to a weak linear programming relaxation as well as an explosion in the number of integer variables and constraints, which might explain why the MILP approach to FMM has not picked up significant interest. A CP formulation is more natural and compact, as we will show in this paper.

\subsection{Classical AI Planning}
Very recently, AI planning techniques were used for FMM~\cite{speck2023finding}. They use a similar state space as AlphaTensor but use various planning tools (with and without exhaustive search) to solve this problem. They compared a number of heuristic and exact planning methods from the literature on matrices of size up to $3\times 3$. However, the experiments show that planning approaches are severely limited, even failing to find Strassen's algorithm for the $2\times 2$ case (see Table 1 in~\cite{speck2023finding}). We will show that our CP approach is significantly more effective as we are able to attack the $3\times 3$ case with $R=23$, matching the known upper bound from the literature.

%%%%%%%%%%%%%%%%%%%%%%%%%%%%%%%%%%%%%%%%%%%%%%%%%%%%%%%%%
\section{Constraint Programming for Fast Matrix Multiplication} \label{sec:methodology}
%%%%%%%%%%%%%%%%%%%%%%%%%%%%%%%%%%%%%%%%%%%%%%%%%%%%%%%%%
In the FMM problem, all variables have the same domain $\mathbb{F}=\{-1,0,1\}$\footnote{One can consider bigger fields such as $\{-2,-1,0,1,2\}$ but the bulk of the work in the literature has been with $\{-1,0,1\}$.}. Since the variable domains are small and this problem is highly structured, CP is a promising solution paradigm. 

% In this section, we will provide a basic CP model, then present the symmetry-breaking constraints that we add to our CP model, the valid inequalities that we introduce. followed by a brief discussion of a decomposition approach.

% \subsection{Constraint Programming Model}
The base CP model for FMM is given in \Cref{CP_model_base}. Let $\mathcal{U}$ denote the set $\{1, \dots, N \cdot M\}$, $\mathcal{V}$ denote the set $\{1, \dots, M \cdot P\}$, $\mathcal{W}$ denote the set $\{1, \dots, N \cdot P\}$, and $\mathcal{R}$ denote the set~$\{1, \dots, R\}$. The CP model uses three sets of variables: $u_{i,r}$ where $i \in \mathcal{U}$, $v_{j,r}$ where $j \in \mathcal{V}$, and $w_{k,r}$ where $k \in \mathcal{W}$; $r \in \mathcal{R}$ in all three cases. Each variable $u_{i,r}$, $v_{j,r}$ and $w_{k,r}$ represents the value of the $i$/$j$/$k^{\mathrm{th}}$ row and $r^{\mathrm{th}}$ column of the matrices $U$, $V$, and $W$. The domain of all variables is $\{-1, 0, 1\}$. The set of constraints presented here requires that the decomposition algorithm's output matches the original tensor multiplication $T_{NMP}$. Therefore the input to the CSP model is 4 integers: $(N,M,P)$ and $R$. The model then reads as:
% \vspace{-5mm}
% \begin{figure}[htbp]
%     \centering
    \begin{align}
        &\sum_{r=1}^\mathcal{R} \left( u_{i,r} \cdot v_{j,r} \cdot w_{k,r} \right) = T_{i,j,k}, & \forall i \in \mathcal{U}, j \in \mathcal{V}, k \in \mathcal{W}\nonumber\\
        & u_{i,r}, v_{j,r}, w_{k,r} \in \{-1,0,1\}, & \forall i \in \mathcal{U}, j \in \mathcal{V}, k \in \mathcal{W}, r \in \mathcal{R} \label{CP_model_base}
    \end{align}
%     \caption{Base CP Model.}
%     \label{CP_model_base}
% \end{figure}
% \vspace{-2mm}

The search space for this (NP-complete) problem grows very quickly with increasing matrix sizes $N,M,P$ and rank $R$. With only one set of equality constraints, a CP solver may struggle with constraint propagation, thus failing to scale with increasing $N,M,P$. To that end, we will introduce additional valid constraints to help CP prune and propagate more efficiently.

\subsection{Symmetry Breaking}
\label{sec:symmetry}
There are many symmetric solutions to the FMM problem. We can reduce the search space of our problem significantly by prohibiting symmetries. 

\subsubsection{Permutation Symmetry}
Since addition is commutative, i.e., $(a_1 + a_2) = (a_2 + a_1)$, there are many equivalent solutions to the tensor decomposition problem. Therefore, any permutation of the columns of matrices $U$, $V$, and $W$ produces an equivalent solution. If we consider Strassen's solution for the $2\times 2$ case, Figure~\ref{fig:permutation} provides an example of two equivalent solutions. 

\newlength\mylenn
\settowidth\mylenn{...}
    
\begin{figure}[htpb]
    \begin{align*}
    \text{sol1: }& \;
       U = \left( \begin{smallarray}{*{7}{wr{\mylenn}}}
  1 & 0 & 1 & 0 & 1 & $-1$ & 0 \\
  0 & 0 & 0 & 0 & 1 & 0 & 1 \\
  0&1&0&0&0&1&0\\
  1&1&0&1&0&0&$-1$
  \end{smallarray} \right)\;\;
        V = \left( \begin{smallarray}{*{7}{wr{\mylenn}}}
  1 & 1 & 0 & $-1$ & 0 & 1 & 0 \\
  0 & 0 & 1 & 0 & 0 & 1 & 0 \\
  0&0&0&1&0&0&1\\
  1&0&$-1$&0&1&0&1
  \end{smallarray} \right) \;\;
        W = \left( \begin{smallarray}{*{7}{wr{\mylenn}}}
  1 & 0 & 0 & 1 & $-1$ & 0 & 1 \\
  0 & 0 & 1 & 0 & 1 & 0 & 0 \\
  0&0&0&1&0&0&0\\
  1&$-1$&1&0&0&1&0
  \end{smallarray} \right) \\
    \text{sol2: }& \;
        U = \left( \begin{smallarray}{*{7}{wr{\mylenn}}}
  $-1$&0&0&0&1&1&1\\
  0&0&0&1&0&0&1\\
  1&0&1&0&0&0&0\\
  0&1&1&$-1$&0&1&0
  \end{smallarray} \right) \;\;
        V = \left( \begin{smallarray}{*{7}{wr{\mylenn}}}
  1&$-1$&1&0&0&1&0\\
  1&0&0&0&1&0&0\\
  0&1&0&1&0&0&0\\
  0&0&0&1&$-1$&1&1
  \end{smallarray} \right) \;\;
        W = \left( \begin{smallarray}{*{7}{wr{\mylenn}}}
  0&1&0&1&0&1&$-1$\\
  0&0&0&0&1&1&1\\
  0&1&1&0&0&0&0\\
  1&0&$-1$&0&1&1&0
  %       U = \left( \begin{smallarray}{p{2.8mm}p{2.8mm}p{2.8mm}p{2.8mm}p{2.8mm}p{2.8mm}p{2.8mm}}
  % 1 & 0 & 1 & 0 & 1 & -1 & 0 \\
  % 0 & 0 & 0 & 0 & 1 & 0 & 1 \\
  % 0&1&0&0&0&1&0\\
  % 1&1&0&1&0&0&-1
  % \end{smallarray} \right)\;\;
  %       V = \left( \begin{smallarray}{p{2.8mm}p{2.8mm}p{2.8mm}p{2.8mm}p{2.8mm}p{2.8mm}p{2.8mm}}
  % 1 & 1 & 0 & -1 & 0 & 1 & 0 \\
  % 0 & 0 & 1 & 0 & 0 & 1 & 0 \\
  % 0&0&0&1&0&0&1\\
  % 1&0&-1&0&1&0&1
  % \end{smallarray} \right) \;\;
  %       W = \left( \begin{smallarray}{p{2.8mm}p{2.8mm}p{2.8mm}p{2.8mm}p{2.8mm}p{2.8mm}p{2.8mm}}
  % 1 & 0 & 0 & 1 & -1 & 0 & 1 \\
  % 0 & 0 & 1 & 0 & 1 & 0 & 0 \\
  % 0&0&0&1&0&0&0\\
  % 1&-1&1&0&0&1&0
  % \end{smallarray} \right) \\
  %   \text{sol2: }& \;
  %       U = \left( \begin{smallarray}{p{2.8mm}p{2.8mm}p{2.8mm}p{2.8mm}p{2.8mm}p{2.8mm}p{2.8mm}}
  % -1&0&0&0&1&1&1\\
  % 0&0&0&1&0&0&1\\
  % 1&0&1&0&0&0&0\\
  % 0&1&1&-1&0&1&0
  % \end{smallarray} \right) \;\;
  %       V = \left( \begin{smallarray}{p{2.8mm}p{2.8mm}p{2.8mm}p{2.8mm}p{2.8mm}p{2.8mm}p{2.8mm}}
  % 1&-1&1&0&0&1&0\\
  % 1&0&0&0&1&0&0\\
  % 0&1&0&1&0&0&0\\
  % 0&0&0&1&-1&1&1
  % \end{smallarray} \right) \;\;
  %       W = \left( \begin{smallarray}{p{2.8mm}p{2.8mm}p{2.8mm}p{2.8mm}p{2.8mm}p{2.8mm}p{2.8mm}}
  % 0&1&0&1&0&1&-1\\
  % 0&0&0&0&1&1&1\\
  % 0&1&1&0&0&0&0\\
  % 1&0&-1&0&1&1&0
  \end{smallarray} \right)
    \end{align*}
    \caption{Two equivalent solutions for Strassen's solution of $2\times 2$ matrix multiplication. sol2 is the $\texttt{lexicographic-strict}$ presentation of this solution.}
    \label{fig:permutation}
\end{figure}

In order to break this symmetry, we introduce a \texttt{lexicographic-strict}\footnote{\url{https://www.ibm.com/docs/en/icos/22.1.0?topic=variables-lexicographic-constraint}} constraint on the $u_{i,r}$ and $v_{j,r}$ variables. When applied to two variable arrays $x$ and $y$, the lexicographic ordering constraint enforces that $x$ is strictly less than $y$ in the defined lexicographic order. Because of the strictness, this also enforces that the two variable arrays must be different. This set of symmetry-breaking constraints is modelled as follows:
\begin{align*}
    &\texttt{lexicographic-strict}([u_{:,r};v_{:,r}], [u_{:,r+1};v_{:, r+1}]), & \forall r \in \mathcal{R}
\end{align*}
where $[u_{:,r};v_{:,r}]$ represents the vector concatenating the $r^{\text{th}}$ column of the matrix $U$ and $V$. In Figure~\ref{fig:permutation}, sol2 satisfies the \texttt{lexicographic-strict} constraint.

\subsubsection{Sign Symmetry}
For the multiplicative $m_i$ terms, one can easily see that multiplying both sets of terms from $A$ and $B$ by $-1$ will result in the same solution. For example, $(a_1 + a_4)(b_1+b_4) = (-a_1 - a_4)(-b_1 - b_4)$, where we could multiply any subset of columns of $U$ and $V$ by $-1$ to achieve the same solution. We call this symmetry the sign symmetry. In order to break it, we introduce the following constraints:
\begin{align*}
    & u_{1,r} \leq 0 & \\
    & u_{i,r} \leq \sum_{i'=1}^{i-1} |u_{i',r}| & \forall r \in \mathcal{R}, i > 1, i \in \mathcal{U}
\end{align*}
The main idea of these constraints is to enforce that the first non-zero entry in a column of~$U$ can only take on the value of $-1$, enforcing that the first entry of the columns is either~0 or~$-1$. The subsequent constraints ensure that for any column $r$, an entry in row $i > 1$ can only be~1 if there has been an entry in the same column in an earlier row with value $-1$. This set of constraints applies to the concatenation of the columns in $U$ and $V$, however, in modelling, it only needs to be applied to the columns of the $U$ matrix as none of the columns can be zero, so the leading $-1$ must appear in the $U$ matrix. By applying these constraints, we make sure that $\{-u_{i,r}\}$ is infeasible for any feasible $\{u_{i,r}\}$. Employing this sign symmetry breaking constraint to sol2 from Figure~\ref{fig:permutation}, we arrive at sol3 shown in Figure~\ref{fig:sign}.

\begin{figure}[htpb]
    \begin{align*}
    \text{sol3: } \;
        U = \left( \begin{smallarray}{*{7}{wr{\mylenn}}}
        % {p{2.8mm}p{2.8mm}p{2.8mm}p{2.8mm}p{2.8mm}p{2.8mm}p{2.8mm}}
  $-1$&0&0&0&$-1$&$-1$&$-1$\\
  0&0&0&$-1$&0&0&$-1$\\
  1&0&$-1$&0&0&0&0\\
  0&$-1$&$-1$&1&0&$-1$&0
  \end{smallarray} \right) \;\;
        V = \left( \begin{smallarray}{*{7}{wr{\mylenn}}}
  1&1&$-1$&0&0&$-1$&0\\
  1&0&0&0&$-1$&0&0\\
  0&$-1$&0&$-1$&0&0&0\\
  0&0&0&$-1$&1&$-1$&$-1$
  \end{smallarray} \right) \;\;
        W = \left( \begin{smallarray}{*{7}{wr{\mylenn}}}
  0&1&0&1&0&1&$-1$\\
  0&0&0&0&1&1&1\\
  0&1&1&0&0&0&0\\
  1&0&$-1$&0&1&1&0
  \end{smallarray} \right) 
    \end{align*}
    \caption{sol3 is the solution derived from enforcing the sign symmetry constraints on sol2.}
    \label{fig:sign}
\end{figure}

Similarly, the same type of sign symmetry-breaking constraints can  be applied to the $W$ factor matrix as follows:
\begin{align*}
    & w_{1,r} \leq 0 & \\
    & w_{k,r} \leq \sum_{k'=1}^{k-1} |w_{k',r}| & \forall r \in \mathcal{R}, k > 1, k \in \mathcal{W}.
\end{align*}
The interpretation is as follows. In Figure~\ref{fig:decomp}, consider $c_4=m_1-m_2+m_3+m_6$, and notice that one can redefine $m_6= (a_3 - a_1)(b_1 + b_2)$ to become $m_6= (a_3 - a_1)(-b_1 - b_2)$ and then rewrite $c_4$ as $c_4=m_1-m_2+m_3-m_6$. Recall that the coefficients of the $m$ terms in an output entry $c_k$ are the entries of row $k$ of factor matrix $W$. The transformation we just performed produces two equivalent solutions and is an instance of ``value symmetry'' that is broken by the above constraint set as it forces the first non-zero entry of a column of $W$ to be $-1$.

\subsection{Valid Inequalities}
\label{sec:valid}
Based on the structure of this problem, we can also introduce a series of valid inequalities that could potentially help a CP solver with propagation.

First, for the $W$ matrix, we know that each multiplicative term $m_r$ must be used at least once for sufficiently small $R$ (i.e., for non-trivial cases of the FMM problem where $R\leq NMP$). This means that the sum of each column in $W$ must be at least one:
\begin{align*}
    & \sum_{k \in \mathcal{W}} |w_{k,r}| \geq 1, & \forall r \in\ \mathcal{R}. 
\end{align*}

Each result term $c_l$ must use at least $M$ terms. This is due to a basic fact in algebraic complexity theory which states that the dot-product of two vectors of size $M$ requires at least $M$ multiplications~\cite{winograd1970number}. This means that the sum of each row of $W$ must be greater or equal to $M$:
\begin{align*}
    & \sum_{r \in \mathcal{R}} |w_{k,r}| \geq M, & \forall k \in \mathcal{W}. 
\end{align*}

Each result term $c_l$ must differ in at least two $m_r$ terms; a simple proof by contradiction is omitted for brevity. This can be modelled as follows:
\begin{align*}
    & \sum_{r \in \mathcal{R}} |w_{k,r}-w_{k',r}| \geq 2, & \forall k \neq k' \in \mathcal{W}.
\end{align*}

Each term in the $A$ and $B$ matrices must appear in at least one of the multiplicative terms $m_r$. This translates to each row of $U$ and $V$ having at least one non-zero term as shown in the constraints below:
%\begin{align*}
%    & \sum_{r \in \mathcal{R}} |u_{i,r}| \geq 1, & \forall i \in \mathcal{U}\\
%    & \sum_{r \in \mathcal{R}} |v_{j,r}| \geq 1, & \forall j \in \mathcal{V}
%\end{align*}
\begin{align*}
    & \sum_{r \in \mathcal{R}} |u_{i,r}| \geq 1,  & \forall i \in \mathcal{U} \\
    & \sum_{r \in \mathcal{R}} |v_{j,r}| \geq 1, & \forall j \in \mathcal{V}.
\end{align*}

Furthermore, each valid product of two terms from the $A$ and $B$ matrices, e.g., $a_2b_3$ for $2\times 2$ matrices, must appear in at least one of the $R$ multiplication terms. For $a_2b_3$ appears in $c_1$ and $c_2$, see Figure~\ref{fig:2by2}. This can be modelled as follows:
\begin{align*}
    & \sum_{r \in \mathcal{R}} |u_{i,r} \cdot v_{j,r}| \geq 1, & \forall \text{ valid }i,j.
\end{align*}

\subsection{Full CP Model}
Finally, the full CP model is presented in Figure~\ref{CP_model_full}. The constraints in~\Cref{cons:tensor_equal} ensure that the output matches the original multiplication tensor and thus the validity of an assignment as a matrix multiplication algorithm. We enforce permutation symmetry-breaking with~\Cref{cons:permutation} and sign symmetry-breaking with~\Cref{cons:sign1,cons:sign2,cons:sign3,cons:sign4}. The valid inequalities are modelled through~\Cref{cons:vi1,cons:vi2,cons:vi3,cons:vi4,cons:vi5,cons:vi6}.

\begin{figure}[ht]
    \begin{align}
        & \sum_{r \in \mathcal{R}} \left( u_{i,r} \cdot v_{j,r} \cdot w_{k,r} \right) = T_{i,j,k}, & \forall i \in \mathcal{U}, j \in \mathcal{V}, k \in \mathcal{W} \label{cons:tensor_equal} \\
        &\texttt{lexicographic-strict}([u_{:,r};v_{:,r}], [u_{:,r+1};v_{:, r+1}]), & \forall r \in \mathcal{R} \label{cons:permutation}\\
        & u_{1,r} \leq 0 & \forall r \in \mathcal{R} \label{cons:sign1}\\
        & u_{i,r} \leq \sum_{i'=1}^{i-1} |u_{i',r}|, & \forall r \in \mathcal{R}, i > 1, i \in \mathcal{U} \label{cons:sign2}\\
        & w_{1,r} \leq 0 & \forall r \in \mathcal{R} \label{cons:sign3}\\
        & w_{k,r} \leq \sum_{k'=1}^{k-1} |w_{k',r}| & \forall r \in \mathcal{R}, k > 1, k \in \mathcal{W} \label{cons:sign4}\\
        & \sum_{k \in \mathcal{W}} |w_{k,r}| \geq 1, & \forall r \in\ \mathcal{R} \label{cons:vi1}\\
        & \sum_{r \in \mathcal{R}} |w_{k,r}| \geq M, & \forall k \in \mathcal{W} \label{cons:vi2}\\
        & \sum_{r \in \mathcal{R}} |w_{k,r}-w_{k',r}| \geq 2,& \forall k \neq k' \in \mathcal{W} \label{cons:vi3} \\
        & \sum_{r \in \mathcal{R}} |u_{i,r}| \geq 1, & \forall i \in \mathcal{U}\label{cons:vi4}\\
        & \sum_{r \in \mathcal{R}} |v_{j,r}| \geq 1, & \forall j \in \mathcal{V} \label{cons:vi5}\\
        & \sum_{r \in \mathcal{R}} |u_{i,r} \cdot v_{j,r}| \geq 1, & \forall \text{ valid }i,j \label{cons:vi6}
    \end{align}
    \caption{Full CP Model with symmetry-breaking constraints and valid inequalities.}
    \label{CP_model_full}
\end{figure}

\subsection{Sparsity-based Problem Decomposition}
Given that the factor matrices that have been found for known decompositions tend to be sparse, we introduce some inexact inequalities to induce sparsity and trim candidate assignments that have a high likelihood to be infeasible or that are unnecessarily dense. For example, observe that Strassen's solution in Figure~\ref{fig:decomp} leads to many zeros in the factor matrices; no $m$ term uses more than 2 out of 4 of the $a$ or $b$ terms, no $c$ term uses more than 4 out of the 7 $m$ terms. It has been observed that as the matrix sizes grow, the best solutions become even sparser.

We first introduce a constraint limiting the number of active (i.e., nonzero) terms in each column $r$ (i.e., multiplication term) of $U$ and $V$. This constraint is written as:
\begin{align*}
    & \sum_{i \in \mathcal{U}} |u_{i,r}|+\sum_{j \in \mathcal{V}} |v_{j,r}| \leq K_1, & \forall r \in\ \mathcal{R}.
\end{align*}
A similar constraint can be imposed on $W$, by restricting that each output must use at most $K_2$ multiplication terms. This constraint is written as:
\begin{align*}
    & \sum_{r \in \mathcal{R}} |w_{k,r}| \leq K_2, & \forall k \in \mathcal{W}. 
\end{align*}
Based on these constraints, $K_1$ has an upper bound of $(NM + MP)$ and $K_2$ is upper bounded by $R$. By observing decompositions for small to medium-scale matrices, we can estimate $K_1$ and $K_2$. For example, for $3\times 3$ matrices with $R=23$, we observe that $K_1 = 9$ and $K_2 = 10$ is the safest estimate possible compared to the upper bounds of $18$ and $23$, respectively, which could restrict the CP search dramatically. Note that one could start with any such estimates of the decomposition parameters $K_1$ and $K_2$, iteratively increasing them if the restricted instances are found to be infeasible by the CP solver, eventually resulting in a complete resolution of the original problem.

\subsection{Cyclic Invariant Formulation}
In contrast to the symmetries of the factor matrices discussed in~\Cref{sec:symmetry}, there exists well-known cyclic symmetry for the multiplication tensors $T_N$ of square matrices. More precisely, it is known that $T_{i,j,k} = T_{j,k,i} = T_{k,i,j}$. The authors in \cite{Benson_2015} proposed to leverage this cyclic symmetry property and parameterize FMM algorithms with cyclic invariant factor matrices: $U = [A B C D], \ V = [A D B C], \ W= [A C D B]$ with  $A \in \{-1,0,1\}^{N^2 \times S}$ and $ B, C, D \in \{-1,0,1\}^{N^2 \times T}$ corresponding to a rank $R = S+3T$. 

Although this parametrization reduces the number of integer variables by a factor of three, helping with the combinatorial nature of the problem, there is no guarantee that the minimal rank decomposition corresponds to solutions that exhibit cyclic symmetry. That being said, Strassen's solution of $R=7$ for $N=2$, which is optimal, exhibits such a symmetry~$(S \in\{1,4\})$, as does the best-known rank of 23 for $N=3$~$(S \in\{2,5,11\})$.  Performing two steps of Strassen's algorithm for $N=2$ yields a rank 49 cyclic invariant solution for $N=4$. It is currently unknown whether a solution of rank less than 49 exists for $N = 4$, let alone one exhibiting cyclic invariance.

We implement Ballard and Benson's cyclic invariant reduction \cite{Benson_2015} of the FMM problem for square matrices by reducing the decision variables of our CP formulation as required and imposing the invariant structure on the factor matrices.

%%%%%%%%%%%%%%%%%%%%%%%%%%%%%%%%%%%%%%%%%%%%%%%%%%%%%%%%%
\section{Experiments} \label{sec:experiments}
%%%%%%%%%%%%%%%%%%%%%%%%%%%%%%%%%%%%%%%%%%%%%%%%%%%%%%%%%
In this section, we present our experimental results starting by showing how our CP approach can recover the best-known upper bounds on the rank in a small amount of time on multiplication problems ranging from the trivial $(N,M,P) = (1,1,1)$ case all the way up to the much harder $(2,2,4)$ and $(3,3,3)$ cases. We then present results for the infeasible cases for $(2,2,2)$. We used IBM's CP Optimizer~(CPO)~22.1.0\footnote{\url{https://www.ibm.com/products/ilog-cplex-optimization-studio/cplex-cp-optimizer}} to solve our CP models. We ran our experiments on a compute cluster of AMD Ryzen Threadripper 2990WX cores with 128 GB of RAM per node.

\subsection{Experimental Setup} \label{sec:hardware}

To ensure the reproducibility and robustness of our results, all our experiments are run with multiple random seeds. This accounts for the often observed performance variability in combinatorial search; this is documented for example in MILP~\cite{lodi2013performance}. To that end, we ran each experiment with 10 different seeds. We assigned 8 cores (CPO's \texttt{Workers} parameter) to the solver for each run (except for more compute-intensive experiments in Section~\ref{sec:cyclic} where we assigned 20 cores) and timed out the experiments after 2 hours. 

% \begin{itemize}
%     \item What values of $K_1$ and $K_2$ did we pick:  $K_1=9$ and $K_2=10$ for $(3, 3, 3)$ with $S=5$ and  $K_1=6$ and $K_2=4$ for $(2, 2, 2)$ with $S=4$.
%     \item Random seed
% \end{itemize}

\subsection{Evaluation Metrics} \label{sec:evaluation_metrics}
We will report the solver runtime and the number of branches during the solution process for completed runs (i.e., runs that returned a feasible solution or a proof of infeasibility). Given that each problem is attempted with multiple random seeds, the shifted geometric mean with a shift of 0.0001\footnote{The shifted geometric mean of a set of $n$ values $t_1,\dots,t_n$ is defined as $\big(\prod_{i=1}^n{[t_i + \text{shift}]}\big)^{\frac{1}{n}}-\text{shift}$. Compared to the arithmetic mean, it is less sensitive to large variations in the values.}, median, minimum, and maximum of the time in seconds and the number of branches will be reported for a complete picture of the results. Runs that terminated due to the time or memory limits will be discussed where applicable. 

\begin{table}[htbp!]
\centering
\small
\caption{Runtime results for the base CP model on various matrix dimensions. ``geo mean'' refers to the shifted geometric mean as described in Section~\ref{sec:evaluation_metrics}; ``med'' refers to the median and ``min''/``max'' to the minimum and maximum, respectively.}
\begin{tabular}{llllll}
\toprule
 &   &   &   & Time (sec) & Num Branches \\
$N$ & $M$ & $P$ & $R$ & geo mean (min, med, max) & geo mean (min, med, max)  \\
\midrule
1 & 1 & 1 & 1 & 0.00 (0.00, 0.00, 0.01) & \num{5.05e+01} (\num{4.50e+01}, \num{5.00e+01}, \num{5.80e+01}) \\
1 & 1 & 2 & 2 & 0.00 (0.00, 0.00, 0.01) & \num{1.31e+02} (\num{1.00e+02}, \num{1.27e+02}, \num{2.13e+02}) \\
1 & 2 & 1 & 2 & 0.00 (0.00, 0.01, 0.01) & \num{1.68e+02} (\num{1.08e+02}, \num{1.69e+02}, \num{2.31e+02}) \\
1 & 1 & 3 & 3 & 0.00 (0.00, 0.01, 0.01) & \num{7.09e+02} (\num{3.47e+02}, \num{7.78e+02}, \num{1.38e+03}) \\
1 & 3 & 1 & 3 & 0.00 (0.00, 0.01, 0.01) & \num{9.44e+02} (\num{3.68e+02}, \num{9.83e+02}, \num{1.82e+03}) \\
1 & 2 & 2 & 4 & 0.01 (0.00, 0.01, 0.01) & \num{4.86e+03} (\num{1.68e+03}, \num{5.27e+03}, \num{7.68e+03}) \\
2 & 1 & 2 & 4 & 0.01 (0.01, 0.01, 0.02) & \num{4.36e+03} (\num{2.63e+03}, \num{4.39e+03}, \num{7.14e+03}) \\
1 & 2 & 3 & 6 & 0.05 (0.02, 0.04, 0.10) & \num{3.51e+04} (\num{1.41e+04}, \num{2.96e+04}, \num{9.83e+04}) \\
1 & 3 & 2 & 6 & 0.05 (0.03, 0.06, 0.12) & \num{3.19e+04} (\num{1.40e+04}, \num{3.19e+04}, \num{7.18e+04}) \\
2 & 1 & 3 & 6 & 0.05 (0.02, 0.04, 0.11) & \num{3.79e+04} (\num{1.45e+04}, \num{3.33e+04}, \num{8.93e+04}) \\
2 & 2 & 2 & 7 & 0.74 (0.28, 0.75, 1.84) & \num{6.41e+05} (\num{2.03e+05}, \num{6.69e+05}, \num{1.70e+06}) \\
1 & 3 & 3 & 9 & 0.37 (0.26, 0.36, 0.60) & \num{2.92e+05} (\num{1.84e+05}, \num{3.09e+05}, \num{4.16e+05}) \\
3 & 1 & 3 & 9 & 0.42 (0.18, 0.52, 0.61) & \num{3.10e+05} (\num{1.64e+05}, \num{3.35e+05}, \num{4.36e+05}) \\
2 & 2 & 3 & 11 & 49.64 (0.98, 71.82, 245.06) & \num{3.40e+07} (\num{6.90e+05}, \num{4.71e+07}, \num{1.74e+08}) \\
2 & 3 & 2 & 11 & 26.47 (6.68, 29.41, 133.29) & \num{1.56e+07} (\num{3.52e+06}, \num{1.39e+07}, \num{9.00e+07}) \\
%2 & 2 & 4 & 14 & \num{7.06e+04} (\num{7.06e+04}, \num{7.06e+04}, \num{7.06e+04}) & \num{2.67e+10} (\num{2.67e+10}, \num{2.67e+10}, \num{2.67e+10}) \\
\bottomrule
\end{tabular}
\label{tab:SAT}
\end{table}

\subsection{Feasible Cases: Searching for Solutions with the Base CP Model}\label{sec:feas}
Table~\ref{tab:SAT} shows the time and number of branches (Num Branches) required by the base CP model (i.e., without symmetry breaking or valid inequalities) to find solution for a range of problems. Our approach was able to find Strassen's solution for the $2\times 2$ matrix multiplication in less than a second whereas the AlphaTensor paper~\cite{fawzi2022discovering} reports a few minutes of model inference to find that solution. 

\noindent\textbf{Performance variability.} In Table~\ref{tab:SAT}, we can see that for $(2,2,3)$ with $R=11$, the worst seed took 245 seconds to find a feasible solution compared to 0.98 seconds for the best seed. This drastic difference in time (and ultimately the number of branches) is an indication that minute parameters such as the seed can significantly impact the CP search. For feasible instances, this phenomenon can be seen as a blessing rather than a curse if one has access to multiple cores: the randomness can be exploited by running multiple copies of the solver, terminating as soon as the first successful run is completed. This has been done in MILP~\cite{fischetti2014exploiting}.

\subsection{Feasible Cases: Sparsity Constraints and Cyclic Invariance Help}
\label{sec:cyclic}
Solving the base CP formulation, with our current time and memory budgets, does not yet yield feasible decompositions for dimensions higher than $(2,2,3)$ or $(2,3,2)$. However, after increasing both the time and memory  limits, we were able to find a solution to the problem for dimension $(2,2,4)$ with $R=14$ in 19.6 hours using the inexact inequalities ($K_1 = 11$ and $K_2 = 7$) developed in~\Cref{sec:valid}. Furthermore, our cyclic invariant formulation (with $S=5$) with inexact inequalities ($K_1 = 9$ and $K_2 = 10$) was able to find a solution for $(3,3,3)$, $R = 23$. More specifically, we ran the cyclic invariant formulation for 10 hours with 5 different seeds and observed that two seeds produced a feasible solution within one hour whereas the other three seeds hit the time limit. Once again, this indicates that performance variability in the CP search is significant for our problem. Additionally, we ran the base CP formulation for $(3,3,3)$ without inexact inequalities for 5 seeds which all hit the time limit of 10 hours, demonstrating the benefit of the reduction of variables for the cyclic invariant formulation and sparsity constraints. We have yet to check whether using only inexact inequalities can help the base formulation for $(3,3,3)$. A similar result was observed for the $(2,2,2)$ case in which the cyclic invariant formulation ($S=4$) with inexact inequalities ($K_1 = 6$ and $K_2 = 4$) produced an average solution time of 0.05 seconds across 10 seeds whereas the base CP model has an average of 1.46 seconds. Our current implementation is not able to find cyclic invariant solutions for $N=4$ with $R = 48$, but we have hope that this approach is a promising tool for the search for new cyclic invariant solutions for square matrix multiplication.

\begin{table}[htbp!]
\centering
\small
\caption{Runtime results for the base CP model and variants to prove infeasibility of $R<7$ for (2,2,2). ``geo mean'' refers to the shifted geometric mean as described in Section~\ref{sec:evaluation_metrics}; ``med'' refers to the median and ``min''/``max'' to the minimum and maximum, respectively. Overall, the use of symmetry-breaking constraints (denoted by the letter ``S'') on top of the base CP formulation (``B'') is crucial for efficient proofs of infeasibility. ``V'' refers to the valid inequalities of Section~\ref{sec:valid} which sometimes complement symmetry-breaking but are not always needed for the fastest results.}
\begin{tabular}{llll}
\toprule
 &  & Time (sec) & Num Branches \\
$R$ & Method &  geo mean (min, med, max) & geo mean (min, med, max)  \\
\midrule
\multirow[t]{4}{*}{1} & B & 0.01 (0.00, 0.01, 0.02) & \num{1.08e+03} (\num{1.06e+03}, \num{1.08e+03}, \num{1.10e+03}) \\
 & B+S & 0.00 (0.00, 0.01, 0.01) & \num{1.00e-04} (\num{0.00e+00}, \num{0.00e+00}, \num{0.00e+00}) \\
 & B+V & 0.00 (0.00, 0.00, 0.01) & \num{1.00e-04} (\num{0.00e+00}, \num{0.00e+00}, \num{0.00e+00}) \\
 & \textbf{B+V+S} & \textbf{0.00} (0.00, 0.00, 0.00) & \num{1.00e-04} (\num{0.00e+00}, \num{0.00e+00}, \num{0.00e+00}) \\
\cline{1-4}
\multirow[t]{4}{*}{2} & B & 0.01 (0.00, 0.01, 0.03) & \num{4.38e+03} (\num{3.72e+03}, \num{4.41e+03}, \num{5.30e+03}) \\
 & B+S & 0.01 (0.01, 0.01, 0.02) & \num{1.08e+03} (\num{1.08e+03}, \num{1.08e+03}, \num{1.08e+03}) \\
 & \textbf{B+V} & \textbf{0.00} (0.00, 0.01, 0.02) & \num{1.33e+03} (\num{1.31e+03}, \num{1.32e+03}, \num{1.34e+03}) \\
 & B+V+S & 0.01 (0.01, 0.01, 0.03) & \num{1.09e+03} (\num{1.07e+03}, \num{1.08e+03}, \num{1.10e+03}) \\
\cline{1-4}
\multirow[t]{4}{*}{3} & B & 0.21 (0.17, 0.21, 0.28) & \num{1.70e+05} (\num{1.42e+05}, \num{1.70e+05}, \num{1.95e+05}) \\
 & B+S & 0.02 (0.01, 0.02, 0.03) & \num{4.13e+03} (\num{3.06e+03}, \num{4.30e+03}, \num{5.32e+03}) \\
 & B+V & 0.20 (0.13, 0.21, 0.25) & \num{1.38e+05} (\num{1.14e+05}, \num{1.35e+05}, \num{1.78e+05}) \\
 & \textbf{B+V+S} & \textbf{0.02} (0.01, 0.02, 0.03) & \num{3.61e+03} (\num{2.52e+03}, \num{3.67e+03}, \num{4.92e+03}) \\
\cline{1-4}
\multirow[t]{4}{*}{4} & B & 43.79 (31.33, 41.89, 66.93) & \num{3.85e+07} (\num{3.06e+07}, \num{3.80e+07}, \num{5.00e+07}) \\
 & \textbf{B+S} & \textbf{0.12} (0.07, 0.13, 0.18) & \num{8.50e+04} (\num{6.94e+04}, \num{8.54e+04}, \num{1.03e+05}) \\
 & B+V & 53.49 (39.10, 48.87, 69.35) & \num{3.70e+07} (\num{3.18e+07}, \num{3.69e+07}, \num{4.23e+07}) \\
 & B+V+S & 0.15 (0.11, 0.15, 0.20) & \num{8.53e+04} (\num{7.34e+04}, \num{8.10e+04}, \num{1.06e+05}) \\
\cline{1-4}
\multirow[t]{4}{*}{5} & B & T.O. (N/A, N/A, N/A) & \num{6.03e+09} (\num{5.56e+09}, \num{5.75e+09}, \num{7.14e+09}) \\
 & B+S & 3.06 (2.28, 3.02, 4.15) & \num{2.22e+06} (\num{1.89e+06}, \num{2.19e+06}, \num{2.67e+06}) \\
 & B+V & T.O. (N/A, N/A, N/A) & \num{5.57e+09} (\num{3.97e+09}, \num{5.83e+09}, \num{6.16e+09}) \\
 & \textbf{B+V+S} & \textbf{2.98} (2.56, 2.94, 3.44) & \num{2.14e+06} (\num{1.91e+06}, \num{2.12e+06}, \num{2.56e+06}) \\
\cline{1-4}
\multirow[t]{4}{*}{6} & B & T.O. (N/A, N/A, N/A) & \num{5.99e+09} (\num{4.53e+09}, \num{5.79e+09}, \num{6.93e+09}) \\
 & \textbf{B+S} & \textbf{429.26} (333.88, 441.63, 528.61) & \num{3.28e+08} (\num{2.94e+08}, \num{3.31e+08}, \num{3.76e+08}) \\
 & B+V & T.O. (N/A, N/A, N/A) & \num{4.67e+09} (\num{3.82e+09}, \num{4.73e+09}, \num{5.48e+09}) \\
 & B+V+S & 517.33 (414.07, 522.81, 640.65) & \num{3.35e+08} (\num{2.97e+08}, \num{3.28e+08}, \num{3.95e+08}) \\
% \cline{1-4}
\bottomrule
\end{tabular}
\label{tab:infeasibility}
\end{table}

\subsection{Infeasible Cases: The Importance of Symmetry Breaking}\label{sec:infeas}
Since CP performs an exhaustive search, it can provide a proof of infeasibility if a given rank $R$ is not achievable for certain matrix dimensions. As expected, the runtime to prove infeasibility significantly increases as we approach the known minimum rank; this can be seen in Table~\ref{tab:infeasibility} for the (2,2,2) case. It is also apparent that the addition of symmetry-breaking constraints helps tremendously when proving infeasibility given that they reduce the search space significantly. More specifically, for $R=6$ in Table~\ref{tab:infeasibility}, it is not even currently possible to prove infeasibility without symmetry-breaking constraints in 2 hours whereas the CP model with symmetry-breaking constraints (B+S) requires around 7 minutes. These results highlight the importance of symmetry-breaking constraints when looking to prove infeasibility.

\section{Conclusion} 
\label{sec:conclusion}
We have proposed a novel CP approach to solve the fast matrix multiplication problem. We have provided a set of constraints for breaking permutation and sign symmetries as well as a set of valid inequality constraints to help CP prune and propagate more efficiently. We provide a decomposition framework that is beneficial for finding feasible solutions for the largest case we have attempted, i.e., $3\times 3$ matrix multiplication. Based on our experimental results, we have been able to solve small instances of this problem within a reasonable amount of time. This is in contrast to some existing search-based approaches (MILP, planning) that seem to struggle. In contrast to the AlphaTensor approach~\cite{fawzi2022discovering}, the CP model is far more natural for this combinatorial task and is uniquely positioned to provide proof of infeasibility for some open problems in this space.

While the results of our approach are promising given the limited amount of computing used, there are several limitations that we aim to address in future work. First, our algorithm struggles to scale for larger matrix dimensions or ranks due to the quick increase in the number of variables of the CP model. Secondly, we have found that the base CP model outperforms the addition of symmetry constraints and valid inequalities in the case of feasible solutions, likely due to the latter's tendency to prune symmetric solutions early in the tree search. However, we believe that our experiment's small matrix dimensions may have skewed these results and valid inequalities may be crucial for larger sizes. Moving forward, we propose several areas for further exploration and improvement:
\begin{itemize}
    \item Conduct larger-scale experiments using larger compute clusters to take advantage of the parallelizability of the CP solver's search procedure.
    \item Analyze the highly structured nature of this problem to develop more valid inequalities that can further reduce the search space of our CP model, including inexact inequalities that may not hold for all matrix multiplication dimensions but help for some cases.
    \item Explore solver parameter tuning, particularly for branching strategies and other important search-related decisions.
    \item Further investigate the idea of sparsity-based problem decomposition as a means of improving the scalability and performance of our approach.
    % \item Consider remodeling this problem as an optimization problem that minimizes the number of multiplication terms used (i.e., minimizing $R$) rather than a CSP
\end{itemize}
%%%%%%%%%%%%%%%%%%%%%%%%%%%%%%%%%%%%%%%%%%%%%%%%%%%%%%%%%

\newpage

\bibliography{refs}

\begin{thebibliography}{10}

\bibitem{Benson_2015}
Austin~R. Benson and Grey Ballard.
\newblock A framework for practical parallel fast matrix multiplication.
\newblock {\em {ACM} {SIGPLAN} Notices}, 50(8):42--53, jan 2015.
\newblock URL: \url{https://doi.org/10.1145\%2F2858788.2688513}, \href
  {https://doi.org/10.1145/2858788.2688513}
  {\path{doi:10.1145/2858788.2688513}}.

\bibitem{blaser2003complexity}
Markus Blaser.
\newblock On the complexity of the multiplication of matrices of small formats.
\newblock {\em Journal of Complexity}, 19(1):43--60, 2003.

\bibitem{gs005}
Markus Bl{\"a}ser.
\newblock {\em Fast Matrix Multiplication}.
\newblock Number~5 in Graduate Surveys. Theory of Computing Library, 2013.
\newblock URL: \url{http://www.theoryofcomputing.org/library.html}, \href
  {https://doi.org/10.4086/toc.gs.2013.005}
  {\path{doi:10.4086/toc.gs.2013.005}}.

\bibitem{brockett1973optimal}
Roger~W Brockett and David Dobkin.
\newblock On the optimal evaluation of a set of bilinear forms.
\newblock In {\em Proceedings of the fifth annual ACM symposium on Theory of
  computing}, pages 88--95, 1973.

\bibitem{de1978varieties}
Hans~F de~Groote.
\newblock On varieties of optimal algorithms for the computation of bilinear
  mappings ii. optimal algorithms for 2$\times$ 2-matrix multiplication.
\newblock {\em Theoretical Computer Science}, 7(2):127--148, 1978.

\bibitem{fawzi2022discovering}
Alhussein Fawzi, Matej Balog, Aja Huang, Thomas Hubert, Bernardino
  Romera-Paredes, Mohammadamin Barekatain, Alexander Novikov, Francisco~J
  R~Ruiz, Julian Schrittwieser, and Grzegorz Swirszcz.
\newblock Discovering faster matrix multiplication algorithms with
  reinforcement learning.
\newblock {\em Nature}, 610(7930):47--53, 2022.

\bibitem{fischetti2014exploiting}
Matteo Fischetti and Michele Monaci.
\newblock Exploiting erraticism in search.
\newblock {\em Operations Research}, 62(1):114--122, 2014.

\bibitem{gleixner2021miplib}
Ambros Gleixner, Gregor Hendel, Gerald Gamrath, Tobias Achterberg, Michael
  Bastubbe, Timo Berthold, Philipp Christophel, Kati Jarck, Thorsten Koch, Jeff
  Linderoth, et~al.
\newblock Miplib 2017: data-driven compilation of the 6th mixed-integer
  programming library.
\newblock {\em Mathematical Programming Computation}, 13(3):443--490, 2021.

\bibitem{lodi2013performance}
Andrea Lodi and Andrea Tramontani.
\newblock Performance variability in mixed-integer programming.
\newblock In {\em Theory driven by influential applications}, pages 1--12.
  INFORMS, 2013.

\bibitem{Smirnov2013TheBC}
Alexey~V. Smirnov.
\newblock The bilinear complexity and practical algorithms for matrix
  multiplication.
\newblock {\em Computational Mathematics and Mathematical Physics}, 53:1781 --
  1795, 2013.

\bibitem{sorber2017mixed}
Laurent Sorber and Marc Van~Barel.
\newblock A mixed-integer linear program formulation for fast matrix
  multiplication, 2017.

\bibitem{speck2023finding}
David Speck, Paul H{\"o}ft, Daniel Gnad, and Jendrik Seipp.
\newblock Finding matrix multiplication algorithms with classical planning.
\newblock In Sven Koenig, Roni Stern, and Mauro Vallati, editors, {\em
  Proceedings of the Thirty-Third International Conference on Automated
  Planning and Scheduling (ICAPS 2023)}. AAAI Press, 2023.

\bibitem{strassen1969gaussian}
Volker Strassen.
\newblock Gaussian elimination is not optimal.
\newblock {\em Numerische mathematik}, 13(4):354--356, 1969.

\bibitem{winograd1970number}
Shmuel Winograd.
\newblock On the number of multiplications necessary to compute certain
  functions.
\newblock {\em Communications on Pure and Applied Mathematics}, 23(2):165--179,
  1970.

\end{thebibliography}

\newpage

\end{document}